\documentclass[AMA,STIX2COL]{WileyNJD-v2}
\usepackage{moreverb}

\newcommand\BibTeX{{\rmfamily B\kern-.05em \textsc{i\kern-.025em b}\kern-.08em
T\kern-.1667em\lower.7ex\hbox{E}\kern-.125emX}}

\articletype{Article Type}%

\received{<day> <Month>, <year>}
\revised{<day> <Month>, <year>}
\accepted{<day> <Month>, <year>}

\begin{document}

\title{Establishing a real-time traffic alarm in the city of Valencia with Deep Learning}

\author[1,2]{Miguel G. Folgado*}

\author[2,3]{Ver\'onica Sanz}

\author[4]{Johannes Hirn}

\author[1]{Edgar Lorenzo-S\'aez}

\author[1]{Javier Uchurreguia}

\authormark{Folgado, Sanz, Hirn,  Lorenzo-Saez, Urchueguia}

\address[1]{\orgdiv{ICTvsCC research group, Instituto Universitario de Tecnologías de la Información y
Comunicaciones (ITACA)}, \orgname{Universidad Politécnica de Valencia, Edificio 8G Ciudad Politécnica de la Innovación}, \orgaddress{Camí de Vera, s/n, Valencia, 46022, Spain}}

\address[2]{\orgdiv{Instituto de Física Corpuscular (IFIC)}, \orgname{Universidad de Valencia-CSIC}, \orgaddress{Carrer del Catedrátic José Beltrán Martinez, 2, Paterna, 46980, Spain}}

\address[3]{\orgdiv{School of Mathematical and Physical Sciences}, \orgname{University of Sussex}, \orgaddress{Sussex House BN1 9RH Falmer, UK}}

\address[4]{\orgdiv{Centro de Investigaciones Sobre Desertificaci\'on (CIDE)}, \orgname{CSIC - Universidad de Valencia - Generalitat Valenciana}, \orgaddress{CV-315, Km 10.7, 46113 Moncada, Valencia, Spain}}

\corres{Email addresses: \email{migarfol@ific.uv.es} (Miguel G. Folgado),
\email{veronica.sanz@uv.es} (Ver\'onica Sanz), \email{johannes.hirn@ext.uv.es} (Johannes
Hirn), \email{edlosae@etsiamn.upv.es} (Edgar Lorenzo-S\'aez), \email{jfurchueguia@fis.upv.es}
(Javier F. Urchueguia)}


\abstract[Abstract]{Urban traffic emissions represent a significant concern due to their detrimental impacts on both public health and the environment. Consequently, decision-makers have flagged their reduction as a crucial goal. In this study, we first analyze the correlation between traffic flux and pollution in the city of Valencia, Spain. Our results demonstrate that traffic has a significant impact on the levels of certain pollutants (especially $\text{NO}_\text{x}$). Secondly, we develop an alarm system to predict if a street is likely to experience unusually high traffic in the next 30 minutes, using an independent three-tier level for each street. To make the predictions, we use traffic data updated every 10 minutes and Long Short-Term Memory (LSTM) neural networks. We trained the LSTM using traffic data from 2018, and tested it using traffic data from 2019.}


\jnlcitation{\cname{%
\author{M. G. Folgado}, 
\author{V. Sanz}, 
\author{J. Hirn}, 
\author{E. Lorenzo-Saez}, and 
\author{J. Urchueguia}} (\cyear{2023}), 
\ctitle{Establishing a Real-Time traffic alarm in the city of Valencia with Deep Learning.}}

\maketitle

\section{Introduction}\label{sec:intro}
Reducing transportation emissions would help reduce climate change as well as enhance urban air quality \cite{paper_clean_cities}, the primary source of greenhouse gas emissions (GHG) at the urban level being indeed road transportation~\cite{EEA2022}. For instance in Valencia, Spain, road transport represents about 60\% of total GHG emissions of the city~\cite{covenant}.

In addition, urban traffic is one of the primary sources of air pollutant in cities \cite{EEA2019}. Given that air pollution continues to be the leading environmental cause of premature death \cite{EuropeanCommision2022a}, this makes it one of the most significant issues faced by cities in the European Union~\cite{Anagnostopoulou}. In particular, poor air quality causes a considerable number of non-communicable diseases such as asthma, lung cancer, and cardiovascular issues, see~\cite{EEA2021}, as well as 300,000 premature deaths annually in the EU alone~ \cite{EuropeanCommision2022a}. On the financial side, the cost of air pollution is projected to range from 231 to 853 billion euros annually~\cite{EuropeanCommision2022a}.

With this in mind, the European Union imposed new limitations for the major air pollutants in Directive 2008/50/EC~\cite{directiva}. The European Commission is currently recommending lowering the maximum allowed concentration for NO$_2$ from 40 g/m$^3$ down to 20 g/m$^3$ and for particulate matter smaller than 2.5 µm (PM2.5) from 25 g/m$^3$ down to 10 g/m$^3$~\cite{EC2022b}. According to EU Directive 2008/50/EC, these limits represent the highest concentrations of each pollutant that may be measured over an extended period of time or over a number of hours by official air quality stations. 

Given these directives, local authorities will have to define new plans for air quality, including in particular short-term measures to reduce pollution and to mitigate the risks if limits are exceeded~\cite{directiva}. Such measures will typically affect the day-to-day operations of significant portions of a city, and as such, can create rejection and distrust among segments of the population, weakening the political structure that provides continuity to pollution-reduction policies.

In this context, we would like to focus on systems geared toward prevention rather than mitigation. To do this, such systems must be capable of warning of a high pollution episode with sufficient spatial and temporal resolution to respond efficiently in the specific areas that would otherwise be reaching the legal pollution threshold.

As a matter of fact, due to the impact of air quality on health, and the importance of traffic on air quality, more and more cities are already implementing or intend to implement low emission zones (LEZs) in the near future. Specifically, more than 325 cities had an LEZ at the end of 2022, while 507 cities planned to implement one by 2025~\cite{Clean_Cities_Campaign}. The objective of LEZs is precisely to prevent highly-polluting vehicles from entering areas of the city, often depending on the current levels of pollution within these areas~\cite{urban_acces}.

However, one major issue with such LEZs is that air quality is typically measured by a small number of official air quality stations located at a few specific points in the city. Yet, Lorenzo-Sáez et al~\cite{Lorenzo-Saez_2021} demonstrated in Valencia that, due to the spatial variability of air quality throughout the city, official air quality stations were unable to detect excess pollution in the most polluted zones. In particular, while none of the seven official stations'  annual averages exceeded the limit of 40 g/m$^3$ for NO$_2$ between 2017 and 2019, a much denser system of 424 passive dosimeter sensors pointed to NO$_2$ annual averages higher than that these limits in 43.7\% of the monitored locations.

This implies that, for LEZs to really protect the population, they need to take into account the geographical variation of pollutant concentration at a more local level. One would thus need to predict (and ideally monitor) pollutant concentrations at a finer scale, i.e. that of a neighborhood or even a street, and enforce traffic restrictions with the same fine resolution in order to avoid any area of the city going above the pollution threshold even while the sparse network of official stations is below threshold. As an added benefit, finer neighborhood-level traffic restriction could also be more acceptable to drivers as they would not block large areas of a city for a long time.

The difficulty with these neighborhood-level traffic restrictions is that they would require a high level of monitoring and an efficient warning system at that same geographical scale. The present work is a first step in this direction, where we build a traffic alarm system for each monitored street. This is based on the LSTM developed in Ref. \cite{folgado2022predicting}, but with a higher time-resolution in order to enable traffic prediction at the level required to implement neighborhood-level traffic restrictions. More specifically, traffic data is provided every 10 minutes, and we forecast the next 30 minutes.

As described in Section~\ref{sec:data}, while we have traffic data for 1,472 road traffic sensors throughout the city, we only have official air quality data from 7 specific locations, and therefore we cannot yet train our network to predict pollution for every single neighborhood. Because of this, we are forced to use a proxy: in Section~\ref{sec:correlation}, we first check that traffic flux and air pollution are indeed correlated when we have  data for both in the same street. This justifies our use of traffic flux as a proxy for air pollution. We then show in Section ~\ref{Sec:pred_algo} that we are able to predict traffic at a fine geographical scale, i.e. in our case for every monitored street segment in the city. We predict traffic to a good accuracy 30 minutes ahead of time, leaving enough time to implement local measures to modify traffic flow at the same granularity as the traffic is predicted and measured. In the future, we hope to implement a denser network of air quality sensors to provide an additional input for our model, and to predict directly this air pollution data instead of the traffic flux, even though the two are heavily correlated.

\section{Data acquisition and description}
\label{sec:data}

The city of Valencia was chosen as pilot case because of the particularly high density of its traffic management system. This system is composed of nearly 3,500 induction loops distributed throughout the city, allowing to monitor 1,472 road segments in a city of 809,501 inhabitants (2023 data) and a metropolitan area of 1.5 million inhabitants. In addition, Valencia city council has a collaboration agreement signed with the Joint Research Centre from the European Commission, in order to use Valencia as City Lab in the framework of the Community of Practice on CITIES~\cite{JRC}.

The alarm system proposed in this work uses the traffic data from 1,472 road segments of Valencia. This information is collected using sensors (induction loops placed below the road surface) which count the number of vehicles driving through that road segment in a given time period (hereafter referred to as traffic flux).

\begin{figure}[h]
\centering
\includegraphics[width = 0.45\textwidth]{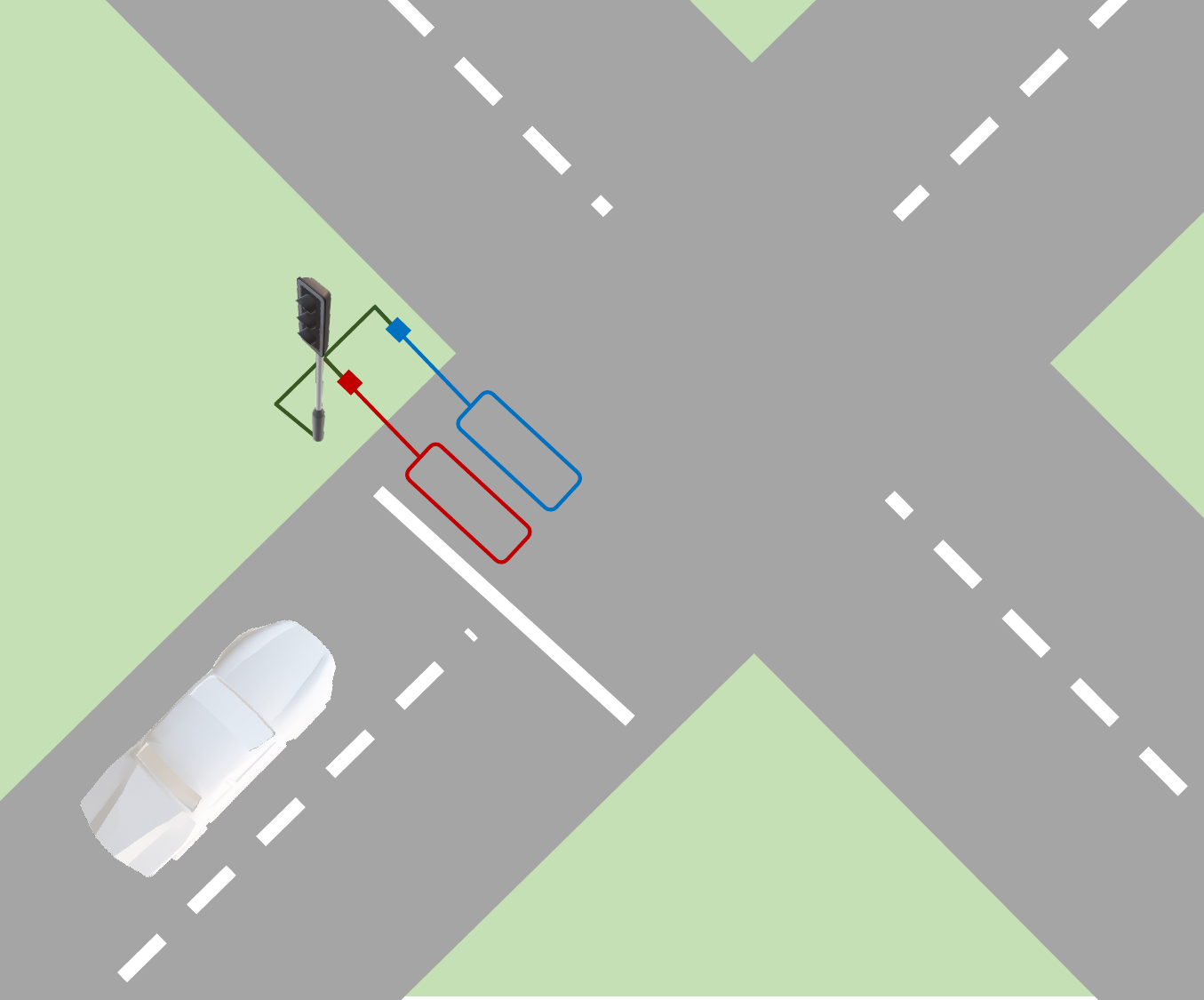}
\caption{Schematics of a sensorized road segment. The red and blue lines are below the road surface. When a vehicle crosses over the electrical circuits, a signal is registered and saved.}
\label{fig:esquema}
\end{figure}

A sensors is depicted schematically in Fig.\ref{fig:esquema}: when a vehicle crosses over the induction loop, the information is sent to a terminal, usually located at the closest traffic light. This terminal is connected to a control room, where information from all induction loops in the city is saved.

In the present study, we have used data aggregated in periods of 10 minutes, whereas in our previous work ~\cite{folgado2022predicting}, we only had access to traffic flux binned at 1-hour granularity.
\begin{figure}[h]
\centering
\includegraphics[width = 0.5\textwidth]{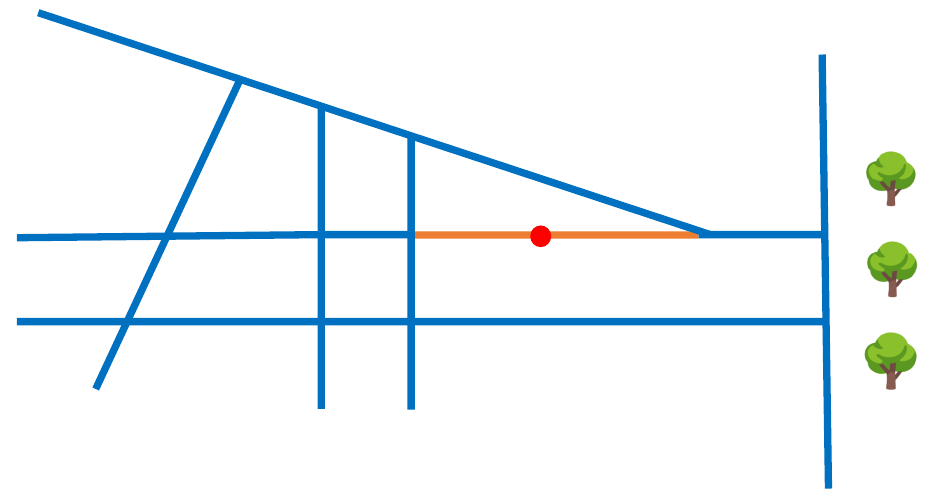}
\caption{Diagram of the road network: the red point represents the location of the induction loop. Roads without sensors are depicted in blue, except the orange line which is the road segment where the traffic flux is assumed to be exactly that given by the induction loop.}
\label{fig:dibujo_sensores}
\end{figure}

Fig.\ref{fig:dibujo_sensores} represents a sensorized road segment: sensor (red point) counts the number of vehicles that drive above it. The sensorized road segment (in orange) is the section of road between the nearest road intersections on both sides: for this sensorized road segment, the traffic flux is equal to that read from the sensor, up to vehicles parking and starting. 

In this study, we have used historical data from 2018 and 2019. Although more recent data is available, we have chosen not to use data form 2020 and onwards since it is affected by government restrictions to mobility in the wake of the Covid-19 pandemic.

\begin{figure*}[!htbp]
\centering
\includegraphics[width = 1.0\textwidth]{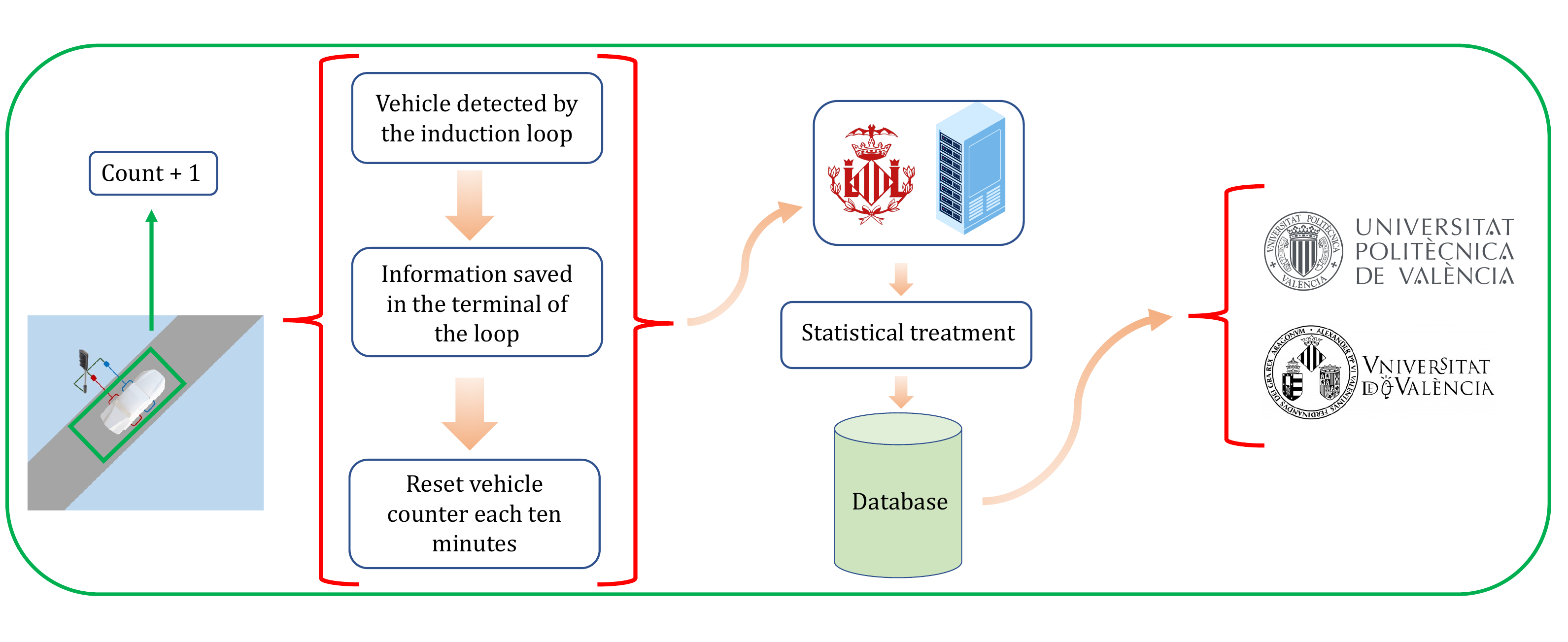}
\caption{The flow of data from the detection of a vehicle to its storage in the computing clusters of the Valencia Universities.}
\label{fig:diagrama}
\end{figure*}

After being collected by the induction loops, the data is sent to servers belonging to the Valencia city hall in charge of the system, see~Fig.\ref{fig:diagrama}. The data is sent in chunks of ten minutes, then filtered to remove low-quality measurements, then saved to a database administered by the city hall. This database is then shared with the two valencian universities (Universidad de Valencia and Universidad Politecnica de Valencia).

Using the above data, we develop an algorithm to predict when traffic flux will reach unusually high levels. This risk is correlated with the emission of different greenhouse gases, such as NO$_\text{x}$. In Sec.\ref{sec:correlation} we analyse and discuss in deep this correlation.
 
\section{Related work} 
\label{sec:related}

There is a large body of research in modelling traffic from a microscopic scale using velocities and densities of vehicles, see for instance Refs.~\cite{art7,PhysRevE.51.1035,RePEc:inm:oropre:v:9:y:1961:i:4:p:545-567}. Here we rely instead on having a large amount of historical data from which to extract patterns using Artificial Intelligence. Specifically, we use the data from 1,472 road sensors in the city of Valencia, Spain, to train a Neural Network (NN), as in our previous work~\cite{folgado2022predicting}, but with a higher time-resolution (10-minute intervals instead of one-hour intervals).

Since our data gives us a traffic flux, i.e. a number of vehicles driving through a location per unit time, this is the variable we must work with, whereas other studies focus on traffic speed~\cite{a8} or travel time~\cite{a36}. Based on the literature (Refs.~\cite{2018arXiv180207007C, cui2019traffic}) and on our previous work~\cite{folgado2022predicting}, we also know that, among the various options for the NN, LSTMs~\cite{2018arXiv181104745M,Song2016DeepTransportPA,cui2018deep,cui2020stacked} are the best candidate for the sparse data we have, as compared to a GraphNN~\cite{2016arXiv160902907K,2013arXiv1312.6203B,2015arXiv151102136A,2016arXiv160601166V,2018arXiv180207007C,lee2022ddp,li2021spatiotemporal,cui2020learning}.

Other examples of NNs that have been employed to predict various features of a traffic network are: Gated Recurrent Units (GRU)~\cite{2014arXiv1406.1078C}, Deep Belief Networks (DBN)~\cite{Huang2014DeepAF,a12}, hybrid models including Convolutional layers~\cite{wang2021hybrid,chen2022spatial,zheng2020hybrid}, stacked autoencoders ~\cite{a2,wei2019autoencoder} and Generative Adversarial Networks (GAN) \cite{2018arXiv180103818L}.

\section{Traffic vs. $\text{NO}_\text{x}$}
\label{sec:correlation}

As mentioned earlier, the main purpose of this paper is to develop a traffic alarm system for the city of Valencia. Early detection of traffic peaks can be used to manage road congestions, which in turn can reduce pollution in certain areas of the city. Currently, it is a well-established fact that air quality deteriorates with increased traffic. In the present Section, we examine the relation between air quality monitoring stations in the city of Valencia and the traffic on nearby roads.

Among the different measurements provided by the air-quality monitoring stations, we focused on $\text{NO}_\text{x}$ in the analysis for two reasons. Firstly, these gases are some of the most harmful gases to human health. Additionally, among the various gases monitored by the station, $\text{NO}_\text{x}$ shows the highest correlation with the traffic data. This is expected since both diesel and gasoline engines are known to emit significant amounts of $\text{NO}_\text{x}$.

The first thing we needed to do in order to study the correlations between traffic and air pollution was to check for any time-lag between the two streams of data. In our case, air quality data is provided in granularity of one hour, so for the current section, we applied the same binning to our traffic data. To find if there is any time-lag between traffic an pollution, we included an offset between the two, and computed the correlation between the two streams of data, depending on this offset.

\begin{figure}[!htbp]
\centering
\includegraphics[width = 0.50\textwidth]{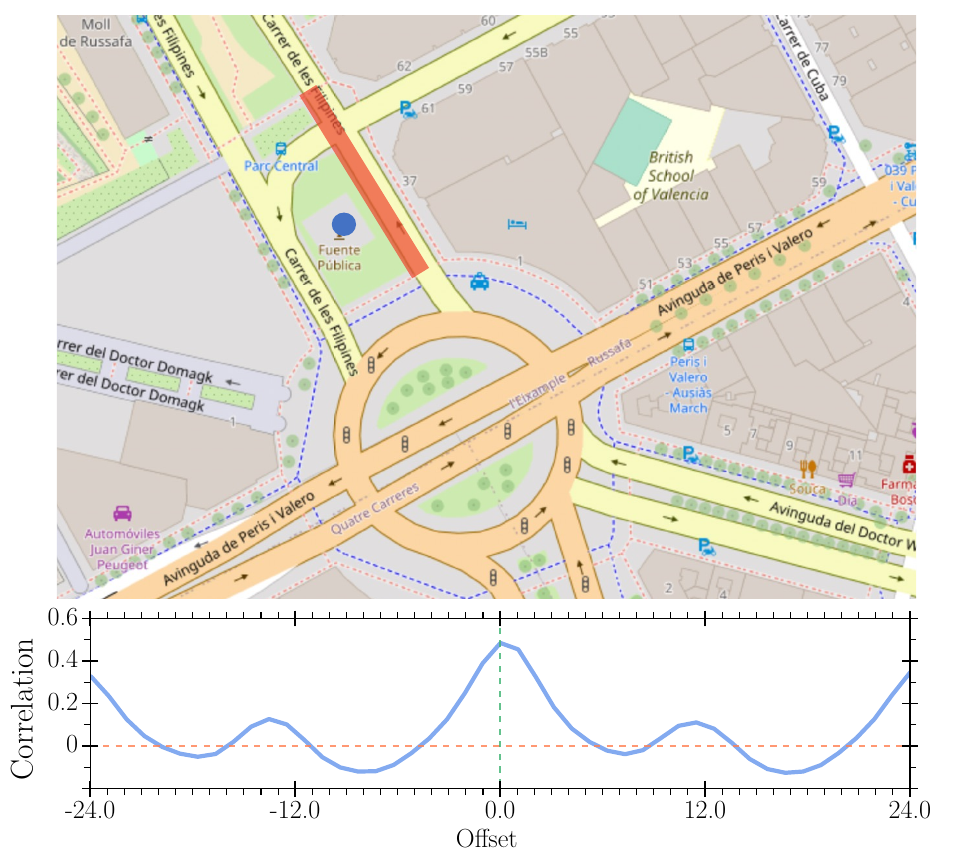}
\caption{Existing correlation between $\text{NO}_\text{x}$ levels and hourly shifted traffic flow in the range of $[-24,24]$. The Pearson correlation coefficient is calculated for the entire month of February 2019. The green-dashed line represents the best offset while the red-dashed line is the zero correlation line.}
\label{fig:offset}
\end{figure}

For the present section, we have focused on a specific air quality monitoring station (marked with a blue dot on the map of Fig.~\ref{fig:offset}) and located near a busy intersection, and in particular next to a road segment for which we have traffic data. In order to study the correlation, we have calculated the Pearson Correlation coefficient:
\begin{equation}
\rho_{X,Y} = \frac{\sum_{i=1}^{n}(X_i - \overline{X})(Y_i - \overline{Y})}{\sqrt{\sum_{i=1}^{n}(X_i - \overline{X})^2} \sqrt{\sum_{i=1}^{n}(Y_i - \overline{Y})^2}},
\label{eq:pearson}
\end{equation}
where $\rho_{X,Y}$ is the Pearson coefficient, and $n$ is the size of the two compared datasets $X$ and $Y$: in our cases $i$ indicates the hour under study, going from $1$ to $n=24 \times 28$ given that February 2019 had 28 days of 24 hours.

The plot in Figure~\ref{fig:offset} represents the correlation for the entire month between  the temporally shifted traffic time series and the unshifted $\text{NO}_\text{x}$ level time series. The results show that the maximum correlation is achieved when the signals are not shifted. This indicates an instantaneous correlation between $\text{NO}_\text{x}$ levels and the traffic flow, suggesting no delay between the two time series. The periodic profile observed in the figure is a result of the inherent periodicity in the traffic data. For example, the traffic flow on Monday at 12:00 PM is highly similar to that on Tuesday at the same time.

\begin{figure*}[h]
\centering
\includegraphics[width = 0.80\textwidth]{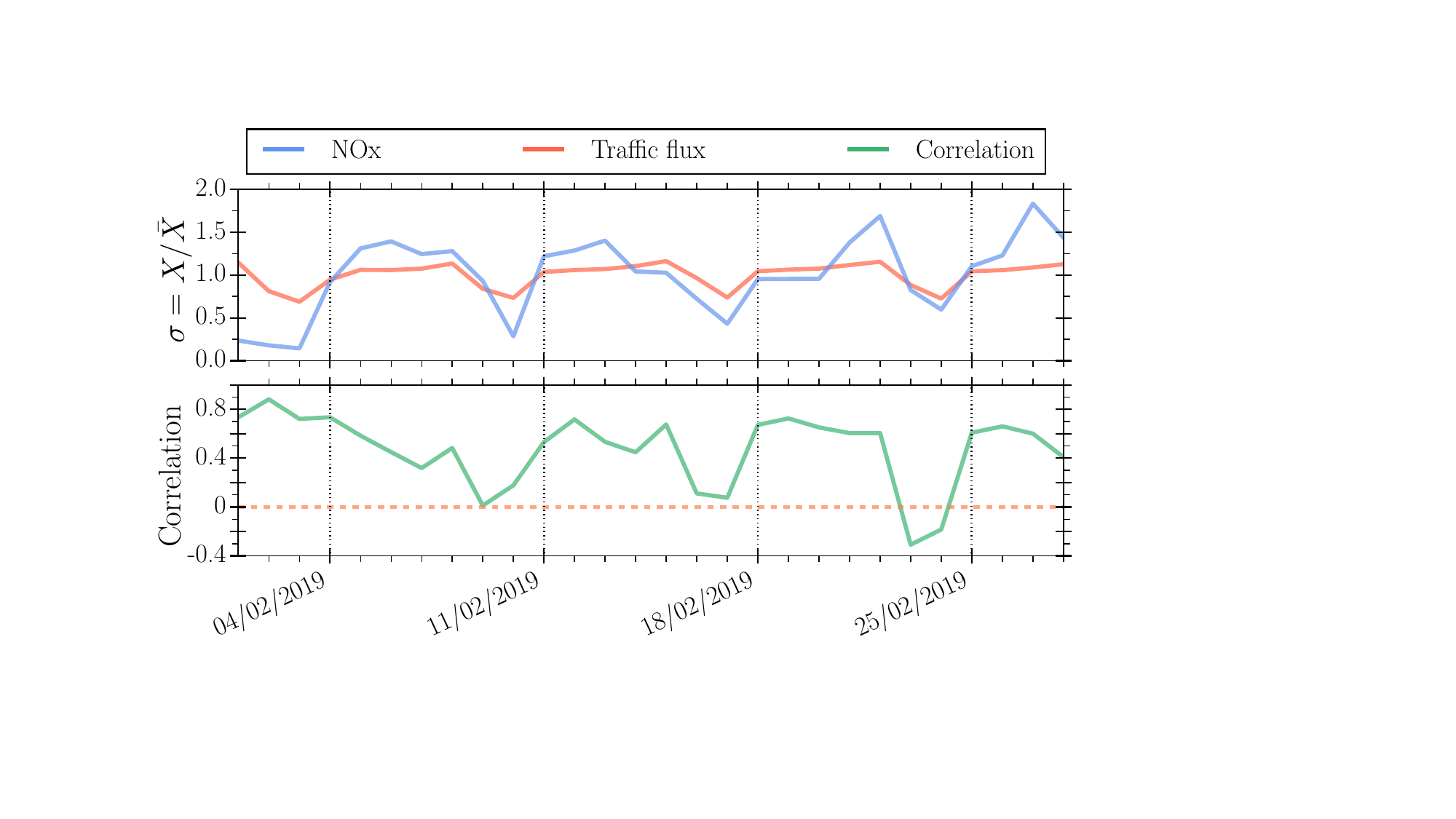}
\caption{Relationship between $\text{NO}_\text{x}$ levels and traffic in the city of Valencia. The data corresponds to the same road segment and station as shown in Fig.\ref{fig:offset}. The top panel shows the variation of daily traffic compared to its monthly average value (red line) and the variation of daily $\text{NO}_\text{x}$ pollution compared to its monthly average value (blue line). In the second panel, the green line represents the Pearson correlation coefficient between the $\text{NO}_\text{x}$ measurements and traffic flow for each day of the month.}
\label{fig:calidad_aire}
\end{figure*}

Once we know that the correlation between traffic flux and $\text{NO}_\text{x}$ levels is instantaneous (at a level of granularity of one-hour), we can ask how this correlation over a month's worth of data breaks down on a day-by-day basis throughout the 28 days of the month, i.e. we can calculate the Pearson correlation of Equation~\ref{eq:pearson} without offset, but with $n=24 \times 1$ instead of $n=24 \times 28$, and to compute that 28 times for each of the 28 days of the month.
The result is shown in Figure~\ref{fig:calidad_aire}, were we display the correlation between the hourly traffic data and hourly $\text{NO}_\text{x}$ levels was calculated for each day of the month of February. The vertical blue-dashed lines in the plot indicate each Monday of the month under study (February 2019). We observe that on Sundays ---the day of the week with the lightest traffic--- the correlation between the two time series decreases. This suggests that the $\text{NO}_\text{x}$ pollution on Sundays is mainly influenced by factors beyond the traffic flow occurring at location and at the same time. In fact, the data indicates the presence of a constant background pollution level, which remains consistent even when the traffic levels drop to almost negligible~\footnote{We are currently working on a more detailed analysis of pollution levels and how various traffic and climatological factors influence them. However, that analysis is beyond the scope of this work.
}.

\section{Methodology}

We describe our methodology in two steps. In~Sec.\ref{Sec:pred_algo}, we describe the improvements on our algorithm of~ Ref.\cite{folgado2022predicting}, used to predict the level of traffic flux for the sensorized road segments. In Sec.\ref{Sec:alarm_sys}, we explain the definition and determination of the alert levels.

\subsection{Predictive algorithm}\label{Sec:pred_algo}
\begin{figure}[h!]
\centering
\includegraphics[width = 0.5\textwidth]{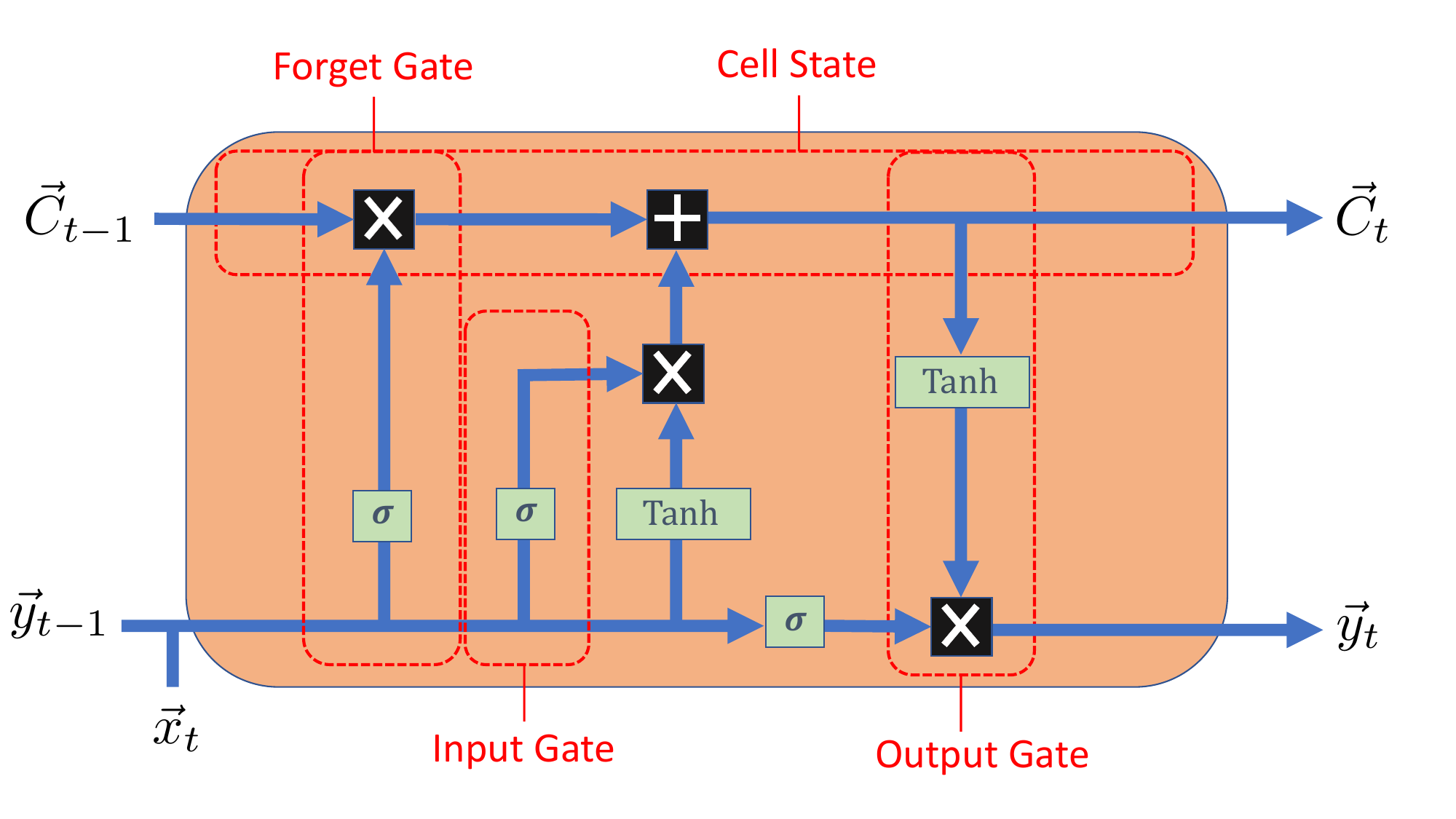}
\caption{Structure of the LSTM used to predict the traffic flux for the city of Valencia, with green rectangles denoting trainable NN layers.}
\label{fig:arquitectura}
\end{figure}

The algorithm we use here to predict the traffic flux for the next 30 minutes is based on Ref.\cite{folgado2022predicting}. In that former study, we only had data with 1-hour granularity. Here we use data with 10-minute granularity in order to improve accuracy. The training data set is the 2018 data, and the prediction has been tested using 2019 data.

We use the standard LSTM architecture as shown in Fig. \ref{fig:arquitectura}. LSTMs have proven to be very effective in forecasting problems, outperforming traditional methods in many cases~\cite{siami2018comparison}. They are particularly well suited for dealing with complex datasets that involve numerous data points in a multidimensional space, which is precisely the case in our study. Specifically, each time slot in our data corresponds to multiple entries (location IDs) and their corresponding traffic flux measurements.

At each time step, the LSTM produces a set of predictions ($\Vec{y}_t$), as depicted in Fig. \ref{fig:arquitectura}. These predictions depend on the features of the current time step ($\Vec{x}_t$), on the previous prediction ($\Vec{y}_{t-1}$), but also on the network's cell state ($\Vec{C}_{t-1}$), a variable designed to retain long-term information from time steps before the previous one. The training process optimizes the parameters of the trainable layers, allowing the LSTM to decide which information to retain and which to forget in order to update its internal cell state $\Vec{C}_{t}$ and generating the prediction $\Vec{y}_t$. As an example, we can consider a specific moment in time, where we want to predict $t$, and a particular street. We will refer to the flow on this specific street as $N^{(1)}$. Thus, the flow we want to predict would be $\vec{y}_t = N^{(1)}_t$. If we are analyzing the prediction using the data from the last 30 minutes (three temporal steps), these values would be stored in the variables:
\begin{align*}
\Vec{y}_{t-1} = \left(N^{(1)}_{t-1}\right) \\
\Vec{y}_{t-2} = \left(N^{(1)}_{t-2}\right) \\
\Vec{y}_{t-3} = \left(N^{(1)}_{t-3}\right)
\end{align*}
Similarly, the features of the neural network would be associated with the flow through the rest of the segments $N^{(n)}$, where $1 \le n \le 1472$. The vectors $\Vec{x}_t$ would take the form:
\begin{align*}
\Vec{x}_{t-1} = \left(N^{(2)}_{t-1}, N^{(3)}_{t-1}, N^{(4)}_{t-1}, ...\right) \\
\Vec{x}_{t-2} = \left(N^{(2)}_{t-2}, N^{(3)}_{t-2}, N^{(4)}_{t-2}, ...\right) \\
\Vec{x}_{t-3} = \left(N^{(2)}_{t-3}, N^{(3)}_{t-3}, N^{(4)}_{t-3}, ...\right). \\
\end{align*}

We trained one LSTM to predict each individual road segment. The training data consisted of the whole 2018 calendar year. The task was to predict three time steps 30-minutes ahead for a single road segment from the previous $6 \times 10$-minute steps for all 1,472 segments. Once trained, each model predicts the traffic 30 minutes into the future for its target road segment in 2.1 milliseconds.

 The hyperparameters used in the training are described in Table~\ref{Tab:tabla_hiper}.
\begin{table}[!htbp]
\centering
\begin{tabular}{c c c }
    \hline
    \bfseries Hyper-parameter & \bfseries Value  \\ 
    \hline
     Validation split & $10\%$ \\
     Learning-rate & $10^{-5}$  \\
     Epochs & $1.1\times 10^4$  \\
     Batch size train & $32$  \\
     Batch size test & $32$  \\
     
\end{tabular}
\caption{Hyper-parameters used in the predictions of each sensorized road segment.}
\label{Tab:tabla_hiper}
\end{table}

\begin{figure*}[!htbp]
\centering
\includegraphics[width = 1\textwidth]{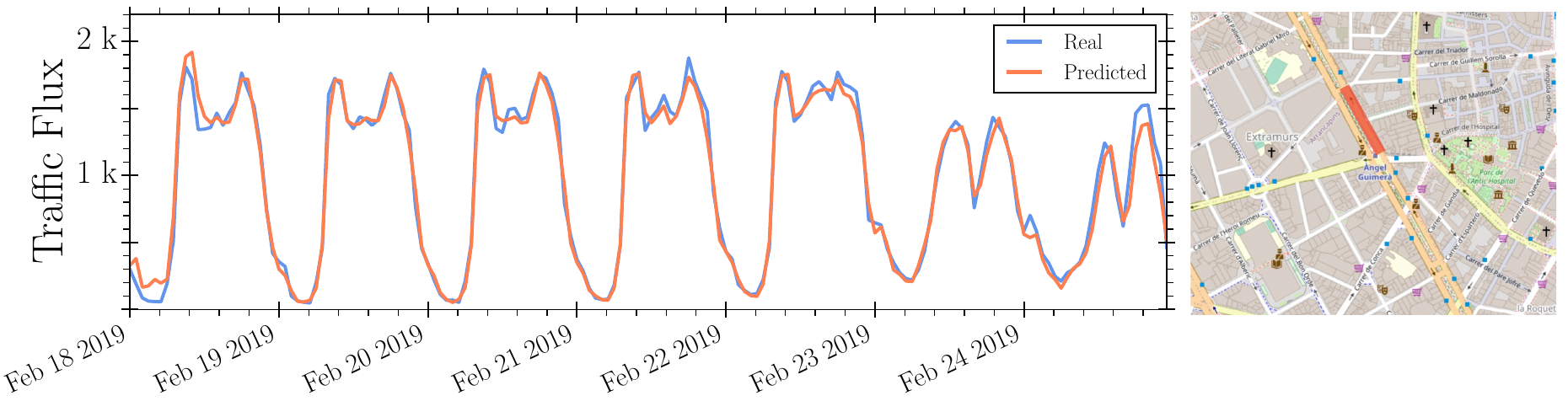}
\caption{Comparison between the real (blue) and the predicted (orange) flux of the sensorized road segment located in Avenida Fernando el Católico, for the selected week.}
\label{fig:resultados_30_mins}
\end{figure*}

The result for an arbitrary road segment, located in the center of the city is shown in Fig.\ref{fig:resultados_30_mins}. The data represented corresponds to the week beginning Monday 18 February of 2019. The blue line is the real traffic flux while the orange line represents the prediction of our LSTM. In this prediction we have used information about the traffic flux one hours before (i.e. 6 steps of 10-minutes) and we have predicted the next 30 minutes. To perform this prediction the algorithm has taken into account the past data from all the sensorized road segments of the city.

As we can see, the prediction is highly accurate: in practice, for the week displayed in Fig.\ref{fig:resultados_30_mins}, the relative error between the prediction and the real data is 13.2 \%.

\subsection{Alarm system}\label{Sec:alarm_sys}

Once we have predicted the traffic flux, it is necessary to establish a methodology to decide whether to raise an alarm for abnormally high traffic for each given road segment. It might then possible to take measures to alleviate traffic flow in the road segment and prevent traffic peaks, thereby avoiding local peaks of air pollution.

\begin{figure}[!htbp]
\centering
\includegraphics[width = 0.50\textwidth]{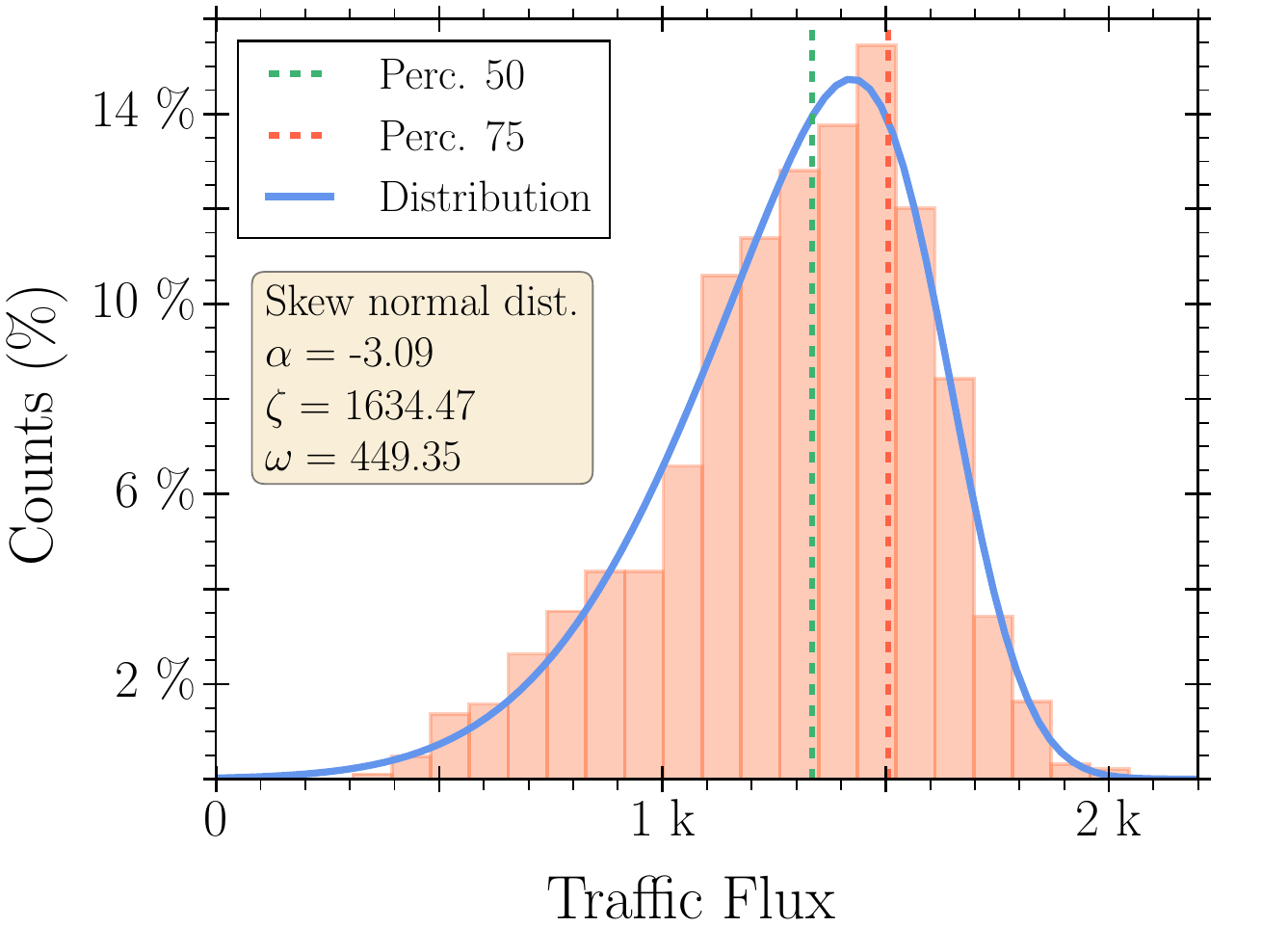}
\caption{Weekly distribution of traffic flow data for a specific road segment (orange). The data is fitted to a Skew Normal distribution (blue). The green and red dotted vertical lines represent the 50th and 75th percentiles, respectively.}
\label{fig:distribucion}
\end{figure}

In Fig.\ref{fig:distribucion}, we can see the distribution of traffic flow data throughout a week on a specific road segment. In general, we have found that most sensorized road segments exhibit a similar data distribution. In all cases, these data can be fitted to a Skew Normal distribution, with probability density function given by:
\begin{equation}
    f(x;\eta,\omega,\alpha) = \frac{2}{\omega} \phi\left(\frac{x-\eta}{\omega}\right) \Phi\left(\alpha\left(\frac{x-\eta}{\omega}\right)\right),
\label{eq:skew}
\end{equation}
where $\phi$ and $\Phi$ are the probability density function and cumulative distribution function of the standard normal distribution, respectively; $\eta$ is the location parameter which determines the shift of the distribution; $\omega$ is the scale parameter which determines the statistical dispersion of the data; and $\alpha$ is the skewness parameter which controls the shape of the distribution. As in the normal distribution, we can define the mean and standard deviation of the distribution:
\begin{equation}
    \mu = \zeta + \omega \delta \sqrt{2/\pi} \; \; \; \; \; \; ; \; \; \; \; \;  \sigma = \omega \sqrt{1 - 2\delta^2/\pi},
\label{eq:skew}
\end{equation}
where $\delta = \alpha/\sqrt{1 + \alpha^2}$.

Using the properties of this distribution could help enhance the sensitivity of the alarm system. However, after analyzing the data, we have found that a percentile-based approach allows for highly reliable anomaly prediction. In this regard, the alarm levels established in our analysis are as follows: low risk below the 50th percentile; medium risk between the 50th and 75th percentiles; and finally high risk above the 75th percentile. In order to extract the percentiles we have computed the traffic flux between 6 AM and 10 PM since January 1st of that year. In the interval between 11 PM and 5 AM the different roads do not have significant flux, as we can see in Fig.\ref{fig:resultados_30_mins}. 

\begin{figure}[!htbp]
\centering
\includegraphics[width = 0.45\textwidth]{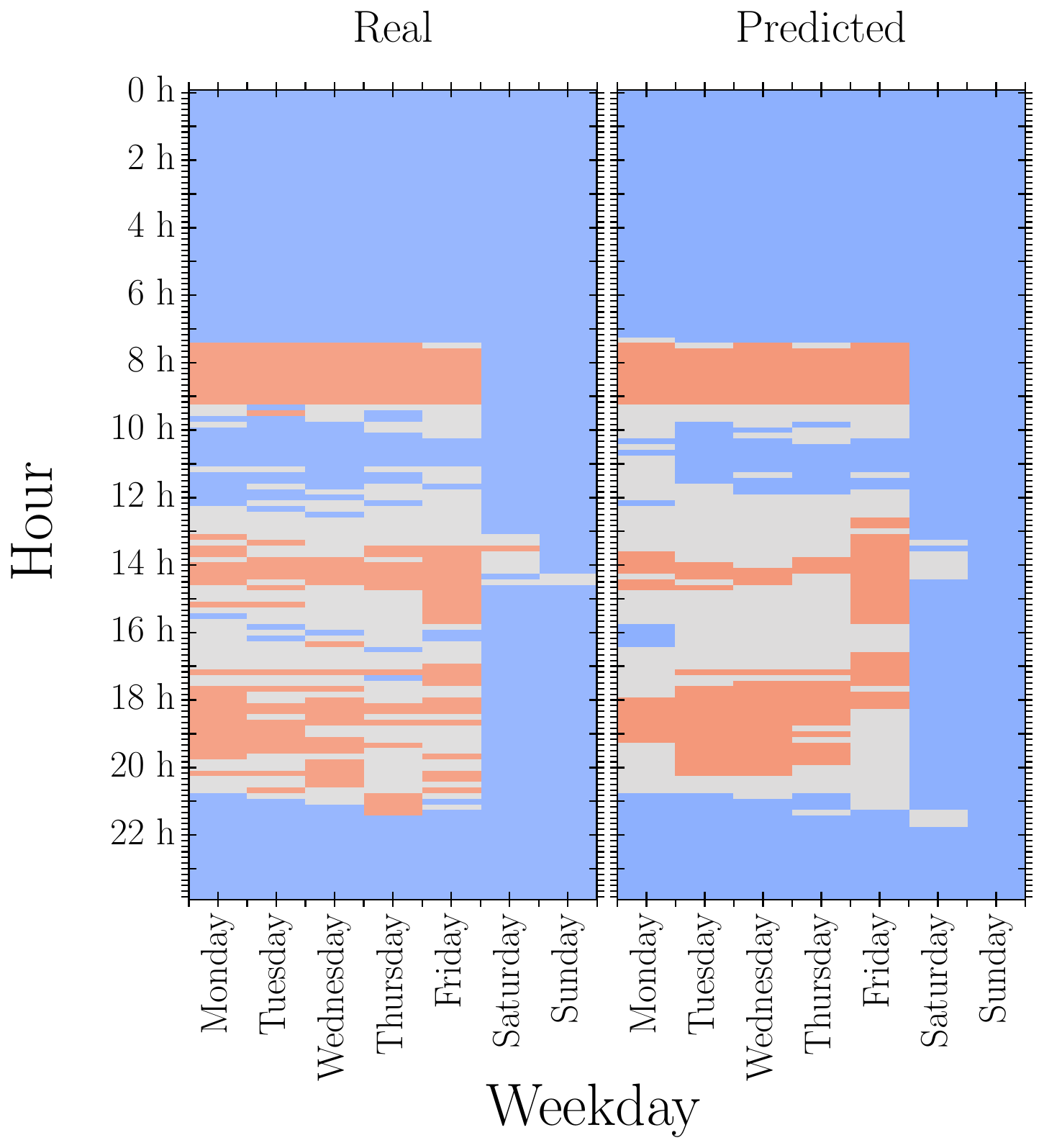}
\caption{Alarm levels of the road segment located in the street Calle Angel Guimera. The left panel represents the real traffic flux while the right panel is the LSTM prediction. Three colours have been chosen in the analysis: blue, for cases where the traffic is under the 50th percentile; grey, when the traffic flux is between the 50th and 75th percentiles and red when the number of vehicles is over the 75th percentile of the distribution. The results are for the selected week.}
\label{fig:comparacion_semanal}
\end{figure}

In Fig.\ref{fig:comparacion_semanal} we have tested our alarm system for the selected week. To make this prediction we have used the data from the past two hours to predict the traffic flux 30 minutes later. The figure represent the traffic flux of a specific road segment in the city of Valencia, in the street Calle Angel Guimera. We use three colors to highlight the status of the traffic flux. When the color is blue, the traffic flux is below the 50th percentile for the chosen road segment. If the color of the square is grey, the traffic flux is between the 50th and 75th percentiles. Finally, the red colour represents cases when the flux of the road segment is above the 75th percentile. The left part of the plot is the real traffic level while the right panel is the prediction of our system. As we can see, in the most part of the week the predictions are highly accurate.

\section{Results}
\begin{figure}[h]
\centering
\includegraphics[width = 0.45\textwidth]{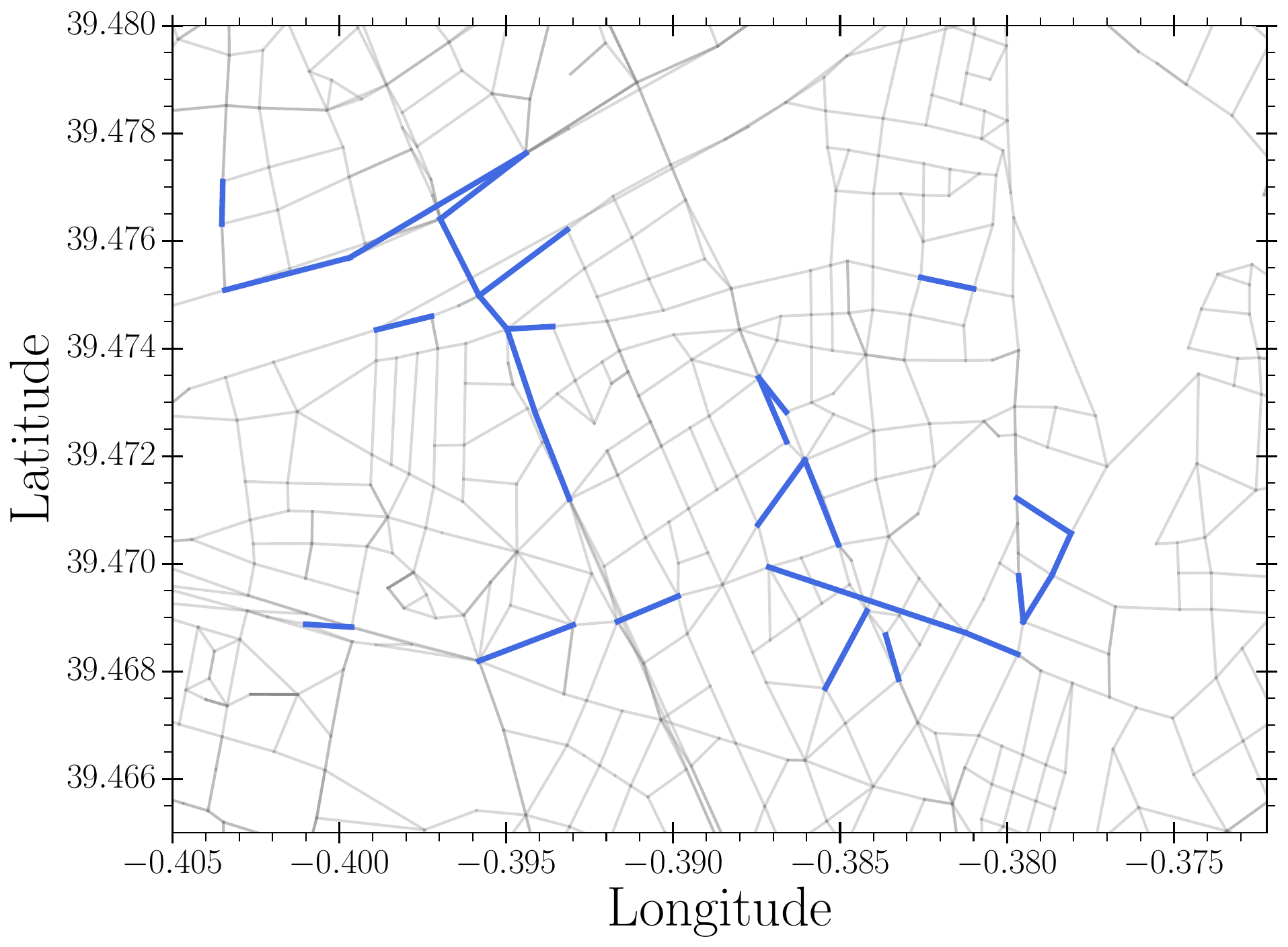}
\caption{Map of the distribution of sensors in a part of the road network of Valencia city. In grey we show the different road segments of Valencia's traffic network. The blue lines represent the road segments for which we have flux data, i.e., the sensorized road segments.}
\label{fig:mapa_tramos_elegidos}
\end{figure}

\begin{figure}[h]
\centering
\includegraphics[width = 0.35\textwidth]{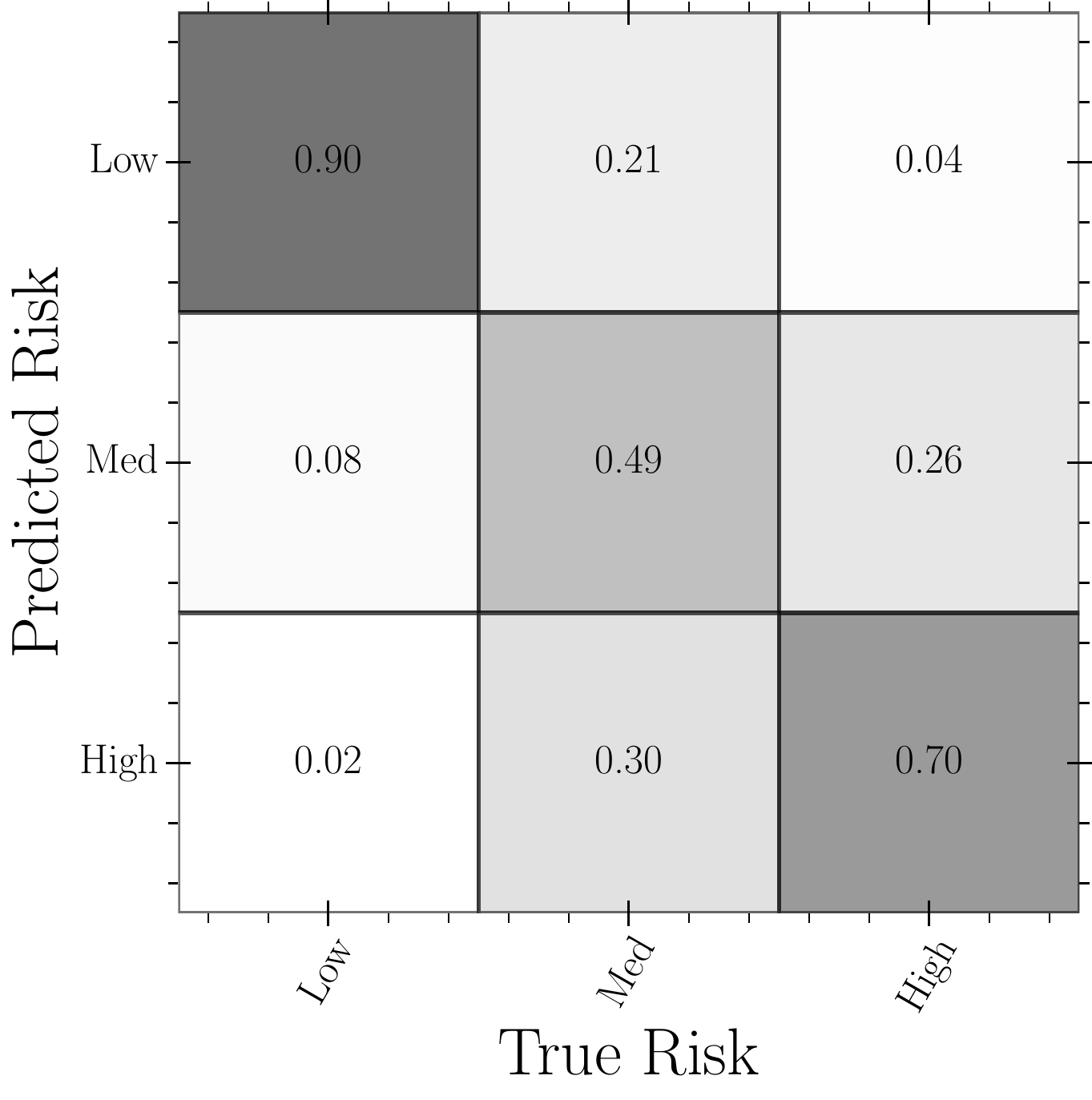}
\caption{Confusion matrix of the Neural Network prediction. The data corresponds to the selected week in the studied area depicted in Fig.~\ref{fig:mapa_tramos_elegidos}.}
\label{fig:confusion}
\end{figure}

In order to test more precisely the system proposed in this work, we have studied a small part of the city, chosen to be representative : it includes part of the old downtown with narrow streets to the East as well as multi-lane avenues to the North-West, fast city exits to the West and finally, intermediate roads in the center and South. In Fig.\ref{fig:mapa_tramos_elegidos} we show the part of the city selected for the analysis: even though we only present an analysis involving that region, we have checked that the results are representative of what happens all over the city.
The results of the prediction for the sensorized road segments located in the area of the city selected in Fig.~\ref{fig:mapa_tramos_elegidos} are shown in Fig.~\ref{fig:confusion}. To build this confusion matrix we have used the same technique as in the Fig.\ref{fig:comparacion_semanal}, but with all road segments in the region depicted in~Fig.\ref{fig:mapa_tramos_elegidos}. As can be seen in the plot, the hours where the risk is low (and traffic is expected to be very fluid) are predicted with high accuracy. In the selected area, the system correctly predicts the low traffic level in the 90\% of the cases. On the other hand, the high levels of traffic are harder to predict: the algorithm succeeds in predicting 70\% of those.
Another interesting conclusion can be extracted from Fig.\ref{fig:confusion}. On the one hand, the errors in all cases mostly occurs with the neighbour risk to the real case. On the other hand, when the real traffic is of medium level, the LSTM tends to overestimate the risk (predicting it to be high more often than to be low). Despite the inaccuracy, we do prefer a model that errs on the side of caution than the other way round.
\begin{figure*}[h]
\centering
\includegraphics[width = 0.8\textwidth]{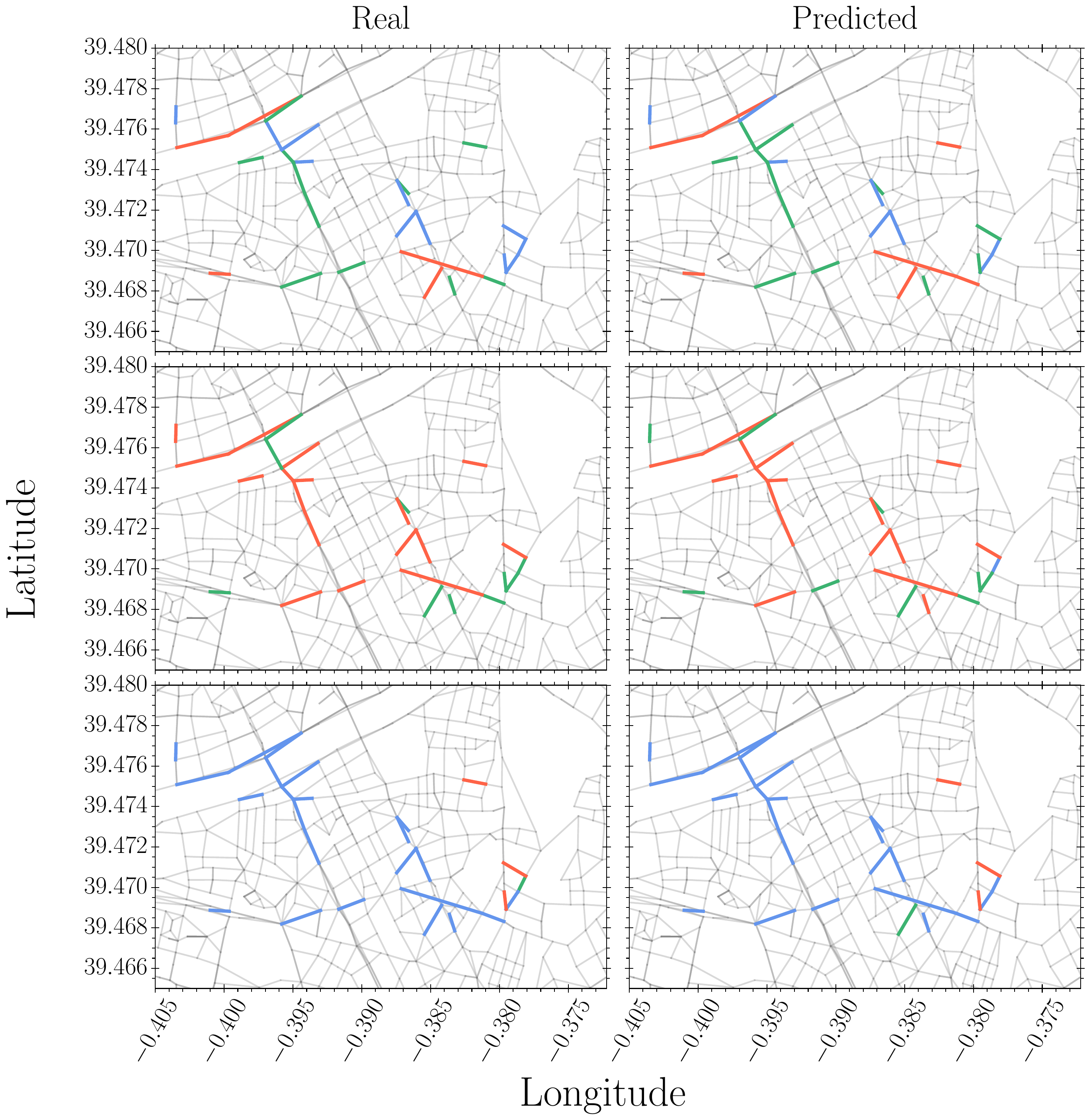}
\caption{Comparison between true and predicted traffic level for our chosen area at different times, specifically form top to bottom: a Tuesday at 2pm, a Friday at 6pm and a Saturday at noon. The blue, green and red colours represents, respectively, the low, medium and high traffic level.}
\label{fig:diferentes_horas}
\end{figure*}

With the methodology described in Sec.\ref{Sec:alarm_sys} we have analyzed the prediction of the traffic level for the next 30 minutes for the sensorized road segments shown in Fig.\ref{fig:mapa_tramos_elegidos}. In Fig.\ref{fig:comparacion_semanal} we have seen the general accuracy of the proposed system. However, we also are interested in the prediction during rush hour, where the traffic flux is higher in general and the forecast of the risk could play an important role. 

In Fig.\ref{fig:diferentes_horas}, we show the prediction of the system for three selected rush hours of the week. The real traffic level is represented by the left panel, while the right panel shows the prediction of the system. In each row of the plot we show a different time of the week. 

In the firs row we can see the traffic flux of a Tuesday at 2pm (February 19 of 2019). In general, during the working days, this hour is one of the most congested of the city: this coincides with lunch break for some jobs and the end of the workday for some others.
The second row shows the traffic flux for Friday at 6pm of the same week. At that time, many families leave the city for their second residence.
Finally, in the last row of the figure, we represent the traffic level for Saturday at noon. The behaviour of the weekend is different to the working days and the congested roads and the rush hours totally change.

\section{Discussion and Outlook}
In this paper we have developed a traffic alarm system based on the traffic data from the city of Valencia during the years 2018, using the past hour to predict the traffic in each road segment 30 minutes in the future 
\begin{figure*}[!htbp]
\centering
\includegraphics[width = 1.0\textwidth]{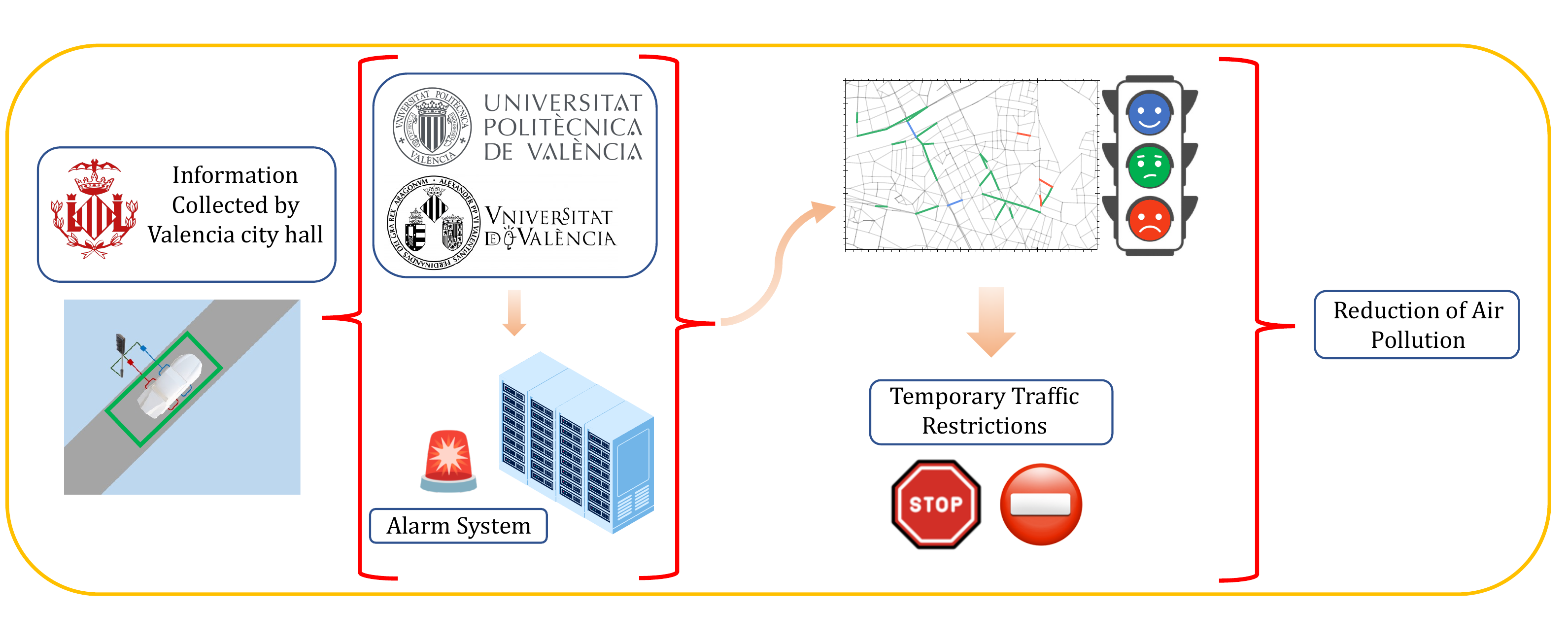}
\caption{Data flow of the implemented software. From data collection by the induction loops to decision-making based on the alarm system results.}
\label{fig:diagrama_conclusiones}
\end{figure*}
The system is based on the Neural Network developed in \cite{folgado2022predicting}, composed, mainly, of an LSTM layer that uses as input the recent traffic data to predict the the future traffic flux. This Neural Network has been trained using the data from the whole year of 2018, and tested on 2019 data. We have designed a system with three alarm levels: low, medium and high, with thresholds adapted to each road segment depending on its own flux distribution. In this sense, the red level of a multi-lane avenue is very different from the red level of a narrow street in the old town.

In order to show the results of the developed system, we have studied a small part of the city. The accuracy of the system in this area is shown in Fig.\ref{fig:confusion}. As we can see, the system predicts the correct alarm level in the most part of the analysed area, with high accuracy.

Our method enables the precise location of zones where traffic, and therefore pollution is about to rise unusually high. Using this information, once can then decide to enforce traffic restriction to avoid such peaks,  as show schematically in the flowchart of Fig. \ref{fig:diagrama_conclusiones}, where we display the flux of the information from road sensor to traffic restrictions. Rather than restricting traffic throughout the entire city or a sizeable portion of it, such short-term actions can be focused on specific streets or neighborhoods. Such interventions with limited scope may be more acceptable to the public.

Additionally, the proposed system might be added as one more variable among others in the optimization of traffic management. In addition, given the geographical resolution and the fact that one can define individual thresholds for each road segment, specific regions can be singled out that are more sensitive to episodes of low air quality (hospitals, schools, nursing homes, etc.), in order to establish a safer limit close to these areas and improve environmental fairness.

In the future, we hope to implement a denser network of IoT air quality sensors that would provide data to further train an LSTM to predict pollution directly instead of its traffic proxy. This would also enable us to set the levels of our alarm systems for each geographical location not only based on its traffic levels, but on the pollution level.

\section{Acknowledgements}

The research of MGF is supported by the \emph{Margarita Salas} postdoctoral fellowship from the \emph{Ministerio de Educación} MS21-038 and the \emph{Generalitat Valenciana} ASFAE/2022/020.
The research of VS is supported by the \emph{Generalitat Valenciana} PROMETEO/2021/083 and the \emph{Ministerio de Ciencia e Innovacion} PID2020-113644GB-I00.
The research of JH is supported by the RESECARIN project TED2021-129682B-I00 from the Spanish \emph{Ministerio de Ciencia e Innovación}.
The research of JFU and ELS is supported by the \emph{Agència Valenciana de la Innovació} (AVI) INNEST/2021/263.

\section{Bibliography}
\bibliography{WileyNJD-AMA}%

\subsection{Appendix Section}

\end{document}